# Protocol effects on feature-based hardware-Trojan detection across Trust-Hub families

Hang Xiao
*Fortinet, Inc.*
Sunnyvale, USA
hangxiao.pisces@gmail.com

Chuhong Xu
*Sony Corporate of America*
San Jose, USA
chuhong.xu@sony.com

Kainan Zhou
*Google LLC*
Mountain View, USA
zhoumark@google.com

Gangzhen Qian*
*Google LLC*
Mountain View, USA
irisqian@google.com

Lu Yi
*Google LLC*
Mountain View, USA
ananbelyi@google.com

*Corresponding author: Gangzhen Qian (irisqian@google.com).

***Abstract*—Trust-Hub reuses host circuits: several files differ mainly in the inserted Trojan. When gates from sibling variants enter both training and test folds, a detector can benefit from host logic it has already seen. We measure that effect instead of proposing another classifier. The corpus contains 49,124 gates from 16 netlists grouped into five host families. We left the parser, 36 gate features, class weighting, model settings, threshold, and family-level aggregation unchanged and altered one choice: the test boundary. The three settings draw test gates from the pooled corpus, withhold a complete netlist, or withhold every variant of one host. The choice matters. Random forest records F1/AP of 0.914/0.978 with pooled gates, 0.636/0.851 with one netlist held out, and 0.460/0.577 with a host family held out. XGBoost falls from 0.946/0.976 to 0.464/0.544 across the same comparison. Logistic regression loses AP, although its fixed-threshold F1 is not monotonic. Each family shows the same pooled-to-family direction. Feature removal, repeated model and simulator seeds, score normalization, parser-related exclusions, and a smaller sample change the size of the gap without reversing it. Aggregation also matters: a gate-weighted average is dominated by the larger ISCAS files, so the headline values give each host family one vote. Bootstrap and jackknife summaries keep the gap positive, but their folds reuse training families. We treat the five family rows as descriptive evidence rather than independent trials. Five host families are too few for a population claim, and the experiment says nothing about transfer to a new cell library or an industrial design. It supports a narrower conclusion: sibling benchmark variants can inflate apparent transfer. Benchmarks with several variants of one host circuit should report family-aware holdouts and all five family results beside pooled scores.***

## I. INTRODUCTION

Hardware Trojans are a persistent concern in integrated-circuit supply chains. They can be introduced during design or fabrication, and their triggers and payloads vary widely. The taxonomy by Tehranipoor and Koushanfar remains a useful account of these insertion points and behaviors [1].

An inserted circuit may remain dormant through ordinary production tests and activate only under a rare internal condition. The survey by Bhunia et al. follows this attack surface through the IC life cycle and reviews countermeasures available at each stage [2]. Early detection is valuable because a late discovery may require costly redesign or physical analysis.

Static netlist analysis offers an earlier view. COTD, for example, derives controllability and observability measures from a gate-level netlist and uses them to locate unusual logic without a golden chip [3]. A learning-based detector works with a similar artifact, although its features and decision rule are learned from labeled designs.

Machine-learning studies usually formulate netlist detection as gate- or signal-level classification. Each instance receives structural, testability, or activity features, followed by a label or score. Trust-Hub complicates the evaluation because it includes several Trojan-inserted versions of the same host design. A random gate split exposes a model to the same circuit on both sides of a fold. Holding out one netlist may still leave a sibling version in training.

We measure the effect of that reuse while keeping the parser, feature matrix, classifiers, weighting rules, and metrics unchanged. Three split schemes are evaluated: pooled gate-level cross-validation, LOCO, and LOFO. Pooled validation permits sibling variants to cross the split; LOCO removes one netlist but may retain a sibling, whereas LOFO removes the complete host family.

## II. PRIOR WORK AND BENCHMARK STRUCTURE

### A. Feature-based detection

Early gate-level learning systems paired hand-designed structural or activity features with conventional classifiers [4]. Later studies explored forests, boosted trees, and graph representations. Dong et al. applied XGBoost to gate-level detection for IoT chips [5]. Those models describe gates differently, but all still depend on which circuits cross the training boundary.

### B. Trust-Hub reuse and the study setting

The experiments assume flattened gate-level netlists, a cell library, and labeled training designs. Classification is performed at cell-instance level from logic visible in the netlist. The evaluation asks a narrow question: how much does prior exposure to sibling host logic change the reported score?

## III. DATASET AND FEATURE PIPELINE

### A. Netlist processing

The parser turns each flattened netlist into a directed graph. Cell instances become nodes, and an edge u → v records a non-clock output of u driving an input of v. Constant ties and simple aliases are resolved before the graph is built. During parsing, the program logs unknown cells, multi-driver nets, and undriven input references.

Salmani et al. developed the design-vulnerability analysis and trust-benchmark methodology that underlies these experiments [6]. The framework supports controlled Trojan insertion and comparison across vulnerable circuit regions.

Shakya et al. cataloged the Trust-Hub circuits and Trojan modifications [7]. The five retained host-family identifiers are defined at the start of Section III.

Trust-Hub groups the retained files by host design. RS232 denotes an RS-232 serial-interface core; s15850, s35932, s38417, and s38584 are standard ISCAS-89 sequential-circuit identifiers. A suffix such as t200 marks a Trojan-inserted variant, so s38584t200 means variant t200 of the s38584 host rather than a variable introduced here.

Three files need documented handling. One payload-stage label in the t300 variant of s35932 is normalized to the spelling used by the same block elsewhere. An undriven input in RS232 variant t1600 and another in s38584 variant t200 are treated as primary-input sources. We exclude the t300 variant of s38584 because duplicated flattened net names permit more than one graph.

The parser stops on unknown cells, residual multi-driver nets, unparsed statements, or undocumented undriven inputs. None remains in the 16 retained files. SHA-256 hashes identify the netlists and the pin-direction library used in the run.

Table I lists the retained sample. Replicated submodules sometimes reuse bus names; surrounding connections resolve most cases, but alternate repairs change the excluded s38584 topology. A sensitivity run later removes every design that needed a repair or unresolved-source assumption.

TABLE I. RETAINED TRUST-HUB BENCHMARKS BY BASE FAMILY

| Family | Circuits | Gates | Trojan gates | Trojan (%) |
|---|---|---|---|---|
| RS232 | 7 | 1,505 | 89 | 5.91 |
| s15850 | 1 | 2,182 | 26 | 1.19 |
| s35932 | 3 | 16,341 | 62 | 0.38 |
| s38417 | 3 | 16,058 | 71 | 0.44 |
| s38584 | 2 | 13,038 | 92 | 0.71 |
| Total | 16 | 49,124 | 340 | 0.69 |

### B. Features and implementation checks

For a gate instance v, let I(v) and O(v) denote its non-clock input nets and output nets, and let G=(V,E) be the directed instance graph. Table II defines all 36 predictors. Hop counts include each reachable instance once and exclude v. Boundary distances are capped at 64, and an unreachable target is also encoded as 64. The 13 structural values are integer graph statistics, the 16 normalized cell-type indicators are binary, and the seven functional summaries lie in [0,1]. No class label is used during feature extraction.

TABLE II. DEFINITIONS OF THE 36 GATE-LEVEL PREDICTORS

| Feature group | Exact definition |
|---|---|
| Structural (13) | fan_in: number of non-clock input pins. fan_out: summed sink-pin counts over output nets. succ2/succ3 and pred2/pred3: distinct other gates reachable within at most two/three forward or reverse hops. dist_pi/dist_po: shortest gate-hop distance from a PI-fed gate/to a PO-driving gate; a boundary-adjacent gate is 0. dist_to_ff/dist_from_ff: shortest directed distance to/from a flip-flop. logic_level: fixed-order logic-level proxy. in_cycle: combinational-cycle flag. sib_max_fanout: largest sink-pin count among input nets. Distances are capped at 64; unreachable is 64. |
| Cell type (16) | One-hot indicators for INV, BUF, AND, NAND, OR, NOR, XOR, XNOR, AO, AOI, OA, OAI, MUX, HADD, ISO/level-shifter, and FF after documented library-name normalization. |
| Functional (7) | out_p1/out_toggle: mean output probability of one/transition rate. out_rare: minimum output rareness r(n)=min[p1(n),1-p1(n)]. in_rare_min/mean and in_toggle_min/mean: minimum and mean rareness/transition rate over non-clock input nets. |

Activity features are estimated with a zero-delay, bit-parallel simulator. Sixteen 64-bit words represent 1,024 trajectories. Four reset cycles and eight warm-up cycles precede 64 measured cycles. Ordinary non-control primary inputs are resampled every cycle with NumPy generator seed 7; scan and test controls are tied low, while active-low resets are asserted during reset and then released. Acyclic logic is evaluated in topological order, and the 53 gates in combinational cycles receive three relaxation sweeps per cycle. For net n, p1(n) is the fraction of one samples over 64 x 1,024 observations, and toggle rate is the fraction of changed samples over 63 x 1,024 adjacent-cycle comparisons.

A separate test harness covers all 66 distinct combinational cell types present in the retained netlists with 850 input vectors and 854 scalar assertions. The cases are repeated with generated Icarus Verilog modules. We also test 11 flip-flop behaviors, including scan, active-low reset, and set/reset combinations, and cross-check ten sequential cases in Icarus. Every listed assertion passes. The harness checks our Boolean implementation and parser interface; vendor timing behavior lies outside this experiment.

## IV. EVALUATION DESIGN

### A. Split protocols

*Pooled cross-validation:* all 49,124 gates are combined and divided into stratified 10-fold cross-validation. Gates from one netlist, and often from one base circuit, can appear on both sides of a fold.

*Leave-one-circuit-out:* one of the 16 netlists is tested in each fold, while the remaining 15 provide training data. Other Trojan variants of the same base circuit may still be present.

*Leave-one-base-family-out:* all variants of one base-circuit family are removed for testing. These five folds give the strongest available separation from reused base logic in the retained Trust-Hub benchmark corpus used here.

Across the three protocols, we reuse the accepted netlists, feature columns, model settings, weighting rule, decision threshold, and aggregation code. Fold membership changes the training size because each protocol removes a different unit. Within this corpus, LOFO gives the cleanest separation from sibling host logic. New libraries and industrial designs are considered separately in Section VII.

### B. *Models and metrics*

The scaler is inside the LR pipeline and is fitted only on each training fold. LR's balanced weights use that complete fold. RF's balanced_subsample recomputes weights for every tree bootstrap, while XGBoost uses the corresponding training-fold negative-to-positive ratio. Thus, the weighting rule is fixed although the realized class ratio varies by fold; no test label enters preprocessing, weighting, or fitting.

For feature vector $x$, LR computes $p_{LR}(x) = \sigma(w^T x+b)$, where $\sigma(z) = (1 + \exp(-z))^{-1}$. RF returns $p_{RF}(x) = T^{-1}\sum_{t=1}^{T} p_t(x)$, and XGBoost returns $p_{XGB}(x) = \sigma(\sum_{m=1}^{M} \eta f_m(x))$. Here $T=M = 400$ and $\eta = 0.1$; all models use $\hat{y}= 1[p(x) \geq 0.5]$.

Table III fixes the training, weighting, and evaluation settings. Unlisted options retain the defaults of the package versions above; no protocol receives separate tuning.

TABLE III. TRAINING, WEIGHTING, AND EVALUATION CONFIGURATION

| Component | Fixed configuration |
|---|---|
| LR preprocessing | StandardScaler(with_mean=True, with_std=True), fitted on each training fold only. |
| Logistic regression | L2 penalty; C=1.0; solver=lbfgs; max_iter=2000; class_weight=balanced. |
| Random forest | 400 trees; Gini; unlimited depth; max_features=sqrt; split/leaf minima 2/1; bootstrap; class_weight=balanced_subsample; random_state=0; n_jobs=2. |
| XGBoost | Binary logistic objective; 400 trees; depth 6; learning rate 0.1; row/column subsampling 0.9; training-fold scale_pos_weight=n_neg/n_pos; tree_method=hist; eval_metric=aucpr; random_state=0; n_jobs=2. |
| Pooled protocol | StratifiedKFold(n_splits=10, shuffle=True, random_state=0). |
| Prediction/ sensitivity | Positive-class probability; threshold 0.5. RF/XGBoost seeds 0-4; simulation seeds 7, 17, and 29. |

The canonical feature build and model runs were executed on an ARM64 host running macOS 26.5.2 and Python 3.12.3. The recorded environment comprises NumPy 1.26.4, pandas 2.1.4, SciPy 1.12.0, scikit-learn 1.4.2, XGBoost 2.0.3, and Matplotlib 3.8.4. RF and XGBoost use two worker threads; LR uses the default lbfgs execution path. The mixed-CV split and primary RF/XGBoost fits use seed 0, while the canonical feature simulation uses seed 7. Model training uses the CPU implementations in these packages; no GPU or distributed backend is enabled in the reported experiments.

The RF and boosted-tree formulations follow Breiman [8] and Chen and Guestrin [9]. Scikit-learn supplies preprocessing, LR, RF, split construction, and metrics [10].

Trojan recall, false-positive rate (FPR), and Trojan-class F1 are evaluated at threshold 0.5, chosen once for all runs. Class weighting affects the score scale, so average precision (AP) is reported alongside these operating-point metrics to summarize score ranking across thresholds.

Each gate receives one out-of-fold score for a given model and protocol. We calculate metrics within each base family and then macro-average the five values. Thus RS232 contributes the same weight as a much larger ISCAS family. In mixed CV and LOCO, a family can collect scores from several fitted models because its gates appear in different test folds.

For the mixed-CV and LOFO comparison, we form five paired family differences. A 200,000-draw family bootstrap, an exact two-sided sign test, and leave-one-family-out jackknife ranges summarize how those observed differences vary. The resampling operates on the five family results and does not refit any classification model.

## V. RESULTS

### A. *Protocol-level performance*

Table IV and Fig. 1 show a larger protocol effect than model effect. RF reaches F1/AP of 0.914/0.978 with pooled gates, 0.636/0.851 with one netlist held out, and 0.460/0.577 with the full base family held out. XGBoost follows the same sequence: 0.946/0.976, 0.732/0.817, and 0.464/0.544. Both tree models therefore lose thresholded F1 and ranking quality once sibling variants no longer cross the split. Because AP falls with F1, the change is not only a threshold artifact. The family and sensitivity results below ask whether one circuit, seed, feature group, or parser decision explains the gap. Later checks repeat the comparison after changing feature groups, seeds, fold scaling, and parser-related exclusions.

TABLE IV. FAMILY-MACRO PERFORMANCE FROM OUT-OF-FOLD SCORES. TPR, FPR, AND F1 USE THRESHOLD 0.5; AP IS THRESHOLD-FREE

| Protocol | Model | TPR | FPR (%) | Trojan F1 | AP |
|---|---|---|---|---|---|
| Mixed (pooled) CV | LR | 0.943 | 8.74 | 0.286 | 0.778 |
| | RF | 0.854 | 0.06 | 0.914 | 0.978 |
| | XGBoost | 0.937 | 0.11 | 0.946 | 0.976 |
| Leave-one-circuit-out (LOCO) | LR | 0.791 | 9.77 | 0.227 | 0.663 |
| | RF | 0.574 | 0.11 | 0.636 | 0.851 |
| | XGBoost | 0.728 | 0.24 | 0.732 | 0.817 |
| Leave-one-base-family-out (LOFO) | LR | 0.692 | 6.13 | 0.246 | 0.609 |
| | RF | 0.391 | 0.31 | 0.460 | 0.577 |
| | XGBoost | 0.549 | 0.74 | 0.464 | 0.544 |

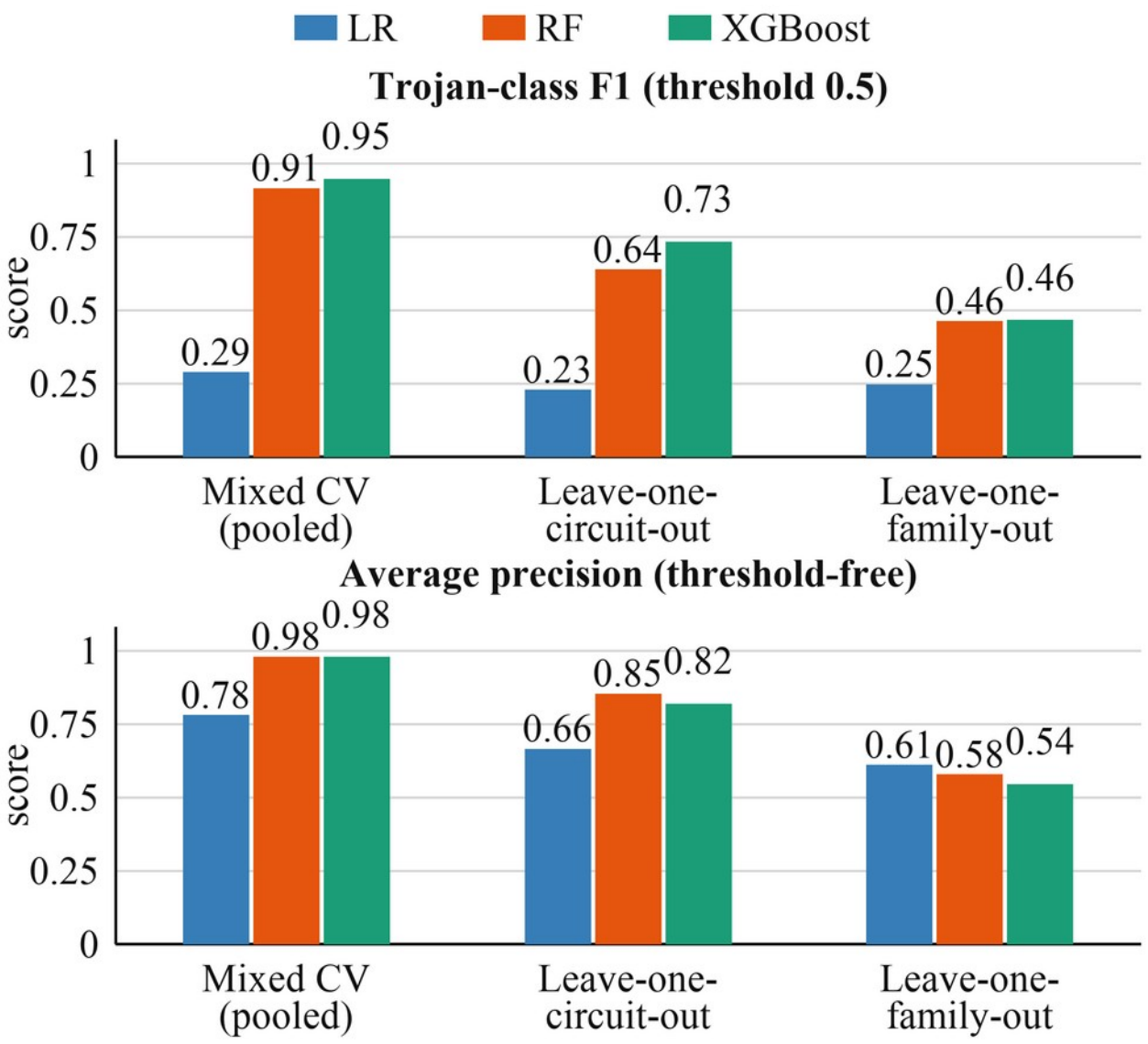

Fig. 1. Family-macro Trojan-class F1 and average precision under the three split protocols.

Across the five families, the pooled-minus-family gap is 0.454 for RF F1 and 0.482 for XGBoost F1; the AP gaps are 0.401 and 0.432. Every paired difference is positive. The exact two-sided sign test gives p=0.0625, so the direction is consistent but the sample is too small for a population claim.

Leave-one-family-out jackknife ranges also stay positive: 0.358-0.549 for RF F1 and 0.416-0.558 for XGBoost. LR is non-monotonic at threshold 0.5, although its AP still declines from 0.778 to 0.609.

## B. Family-level variation

Fig. 2 reports recall for each held-out host family. It is highest for the single s15850 design (0.92-1.00) and lowest for s38584 (0.11-0.20). The smaller RS232 family yields 0.07 for RF, 0.29 for XGBoost, and 0.63 for LR. Hasegawa et al. [11] selected an 11-feature subset; here the feature matrix stays fixed while the split changes.

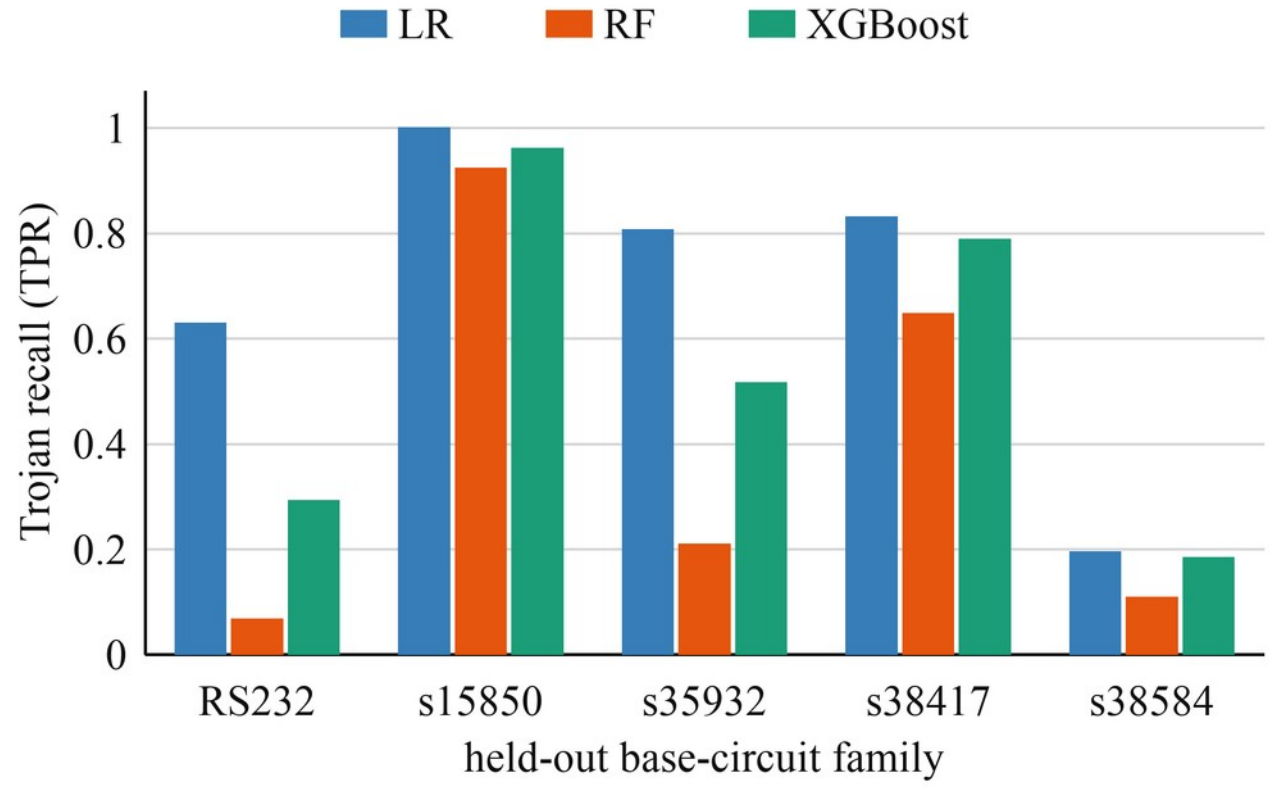

Fig. 2. Trojan recall by held-out base-circuit family under LOFO.

Four families contain more than one Trojan variant. Among them, 46,626 of 46,628 benign gate instances share a name with an instance in a sibling variant. This near-duplicate host logic explains why pooled and circuit-level splits can look stronger than a family holdout.

An identically named gate does not guarantee an unchanged neighborhood, and a model may combine several correlated cues. Kurihara et al. [12] derive trigger-oriented structural features for gate-level detection. LR provides a useful check here: its thresholded F1 is non-monotonic even as AP declines.

Family-macro averaging gives every host family equal weight. Gate-weighted scores are dominated by the larger ISCAS designs; the small RS232 designs create many circuit folds but comparatively few gates. For the single s15850 design, the circuit and family holdouts use the same test file.

## C. Feature and sensitivity checks

Table V repeats the evaluation with restricted feature sets. Removing cell-type indicators changes RF little and lowers XGBoost under LOFO. Structural features retain the pooled-to-LOFO ordering for both tree models, while the seven activity features perform poorly under every split.

TABLE V. FIXED-SAMPLE FEATURE ABLATION. ENTRIES ARE FAMILY-MACRO F1 FOR MIXED/LOCO/LOFO AT THRESHOLD 0.5

| Feature set | RF | XGBoost |
|---|---|---|
| Full feature set (36) | 0.914 / 0.636 / 0.460 | 0.946 / 0.732 / 0.464 |
| No cell-type indicators (20) | 0.904 / 0.640 / 0.461 | 0.926 / 0.684 / 0.407 |
| Structural only (13) | 0.920 / 0.736 / 0.544 | 0.945 / 0.752 / 0.439 |
| Functional only (7) | 0.215 / 0.128 / 0.064 | 0.226 / 0.128 / 0.058 |

One feature identifies all 30 half-adders as Trojan gates, and all occur in a single s38584 variant. Removing it changes RF F1 by at most 0.006 and XGBoost F1 by at most 0.016; it does not inflate one protocol consistently.

Negishi et al. combine boosted-tree predictions with probability propagation through the circuit [13]. Our models do not include that propagation step; the column-removal check simply tests whether one cell type dominates their predictions.

Across activity seeds 7, 17, and 29, XGBoost keeps the same protocol order for both metrics. The largest within-protocol standard deviation is 0.011 for F1 and 0.007 for AP. Mixed-to-LOFO gaps remain 0.482–0.500 for F1 and 0.420– 0.432 for AP. Only the simulation initialization changes during this sensitivity run alone.

Model seeds 0–4 yield the same ordering in all 20 model-metric-seed comparisons. The largest standard deviation is 0.020 for F1 and 0.010 for AP. LOFO F1 ranges from 0.443 to 0.469 for RF and from 0.419 to 0.473 for XGBoost. Repeating seed 0 in the pinned environment reproduces the stored scores within numerical precision here.

We next replace each held-out score with its tie-preserving percentile within that fold and recompute family AP. RF becomes 0.978/0.807/0.577 and XGBoost becomes 0.978/0.755/0.544. LOCO changes most, indicating greater sensitivity to score scales from different fits. The protocol order and the mixed-to-LOFO gap remain.

Finally, removing the three designs with a documented repair or unresolved-input assumption leaves 13 designs, 36,892 gates, 211 positives, and all five families. RF F1 becomes 0.680/0.523/0.358 for mixed/LOCO/LOFO, and XGBoost becomes 0.892/0.607/0.429. Both tree models retain the downward order on this reduced sample.

The reduced sample changes the pooled RF result more than the corresponding XGBoost result, so the size of the gap depends on both model and sample. Across these changes, however, the protocol ordering is stable. It appears in the full feature matrix, in the structural-only matrix, and after the designs with parser assumptions are removed. These runs use different numbers of gates and positives, yet none makes the family holdout easier than either reuse-permitting protocol tested in the present study.

## VI. Discussion

### A. Reading pooled and held-out scores

Gate weighting and family weighting answer different questions. A gate-weighted score is dominated by the larger ISCAS designs; a family-macro score gives RS232 and s15850 the same influence as s38584. Table IV uses family-macro values, while the stored out-of-fold scores permit either aggregation without refitting.

LOCO is intermediate only for the four families with more than one retained variant. For s15850, which contributes one netlist, LOCO and LOFO are the same fold. For RS232, s35932, s38417, and s38584, LOCO removes the test file but leaves sibling host logic in training. It therefore measures transfer between variants of a familiar design. LOFO removes that shortcut.

The feature matrix makes this distinction concrete. Fan-in, local cones, logic level, cell type, and distances tend to survive Trojan insertion. Trees can reuse those combinations without memorizing an instance name. Once a family is withheld, gates from a new topology must be placed using partitions learned from other hosts. AP falls because many Trojan gates move below benign gates, not merely because 0.5 is a poor threshold. Structural-only runs retain the gap; activity-only features do not provide a competitive detector.

A compact report should put pooled and family-held-out columns side by side and expose the five family rows. It should also state the threshold, averaging unit, and whether sibling variants cross the split. GNN4Gate [14] learns over the directed netlist, but a learned representation does not remove the need to define the training boundary.

### B. Operational interpretation

Deployment adds a separate calibration problem. The threshold must reflect review capacity, false-positive tolerance, and the target synthesis flow. Trust-Hub does not determine those quantities.

Threshold choice and transfer are separate decisions. Lowering the threshold may recover recall on a held-out family but cannot restore AP ranking. A deployment study therefore needs calibration data from its target domain and a split matching the novelty expected after release.

Calibration data should come from the intended library and synthesis flow, with the hierarchy available at inference time. A pooled split can be appropriate when future variants belong to a known host family; a family holdout is the relevant stress test when the host is new.

The pooled split and the family holdout also imply different deployment settings. The first screens another Trojan variant of a known host. The second tests a host whose logic has not entered training. Neither score substitutes for the other.

GNN4TJ [15] predicts at design level from RTL data-flow graphs. Its labels and representation differ from our gate-level setup, so the absolute scores are not directly comparable. The common question is whether related host families cross the training boundary.

TrojanSAINT [16] samples gate-level subgraphs for localization. Running it with the same family boundary would separate gains from graph sampling and representation learning from gains caused by familiar host logic.

## VII. Threats to Validity

### A. Construct validity

Trust-Hub names define the positive class. Trigger, payload, and associated inserted logic are all counted as Trojan gates. A narrower label boundary would change both prevalence and the meaning of a gate-level error. We keep the 0.5 operating point fixed across runs and report AP to separate ranking from threshold choice.

Base families follow the named Trust-Hub hosts. That grouping removes the clearest form of reuse, but it cannot identify common synthesis scripts, library cells, or repeated IP blocks that the benchmark metadata do not record.

### B. Internal validity

The parser decisions are explicit. One signal follows the repeated payload-stage pattern; two undriven references are treated as primary-input-like sources; and the t300 variant of the s38584 host is excluded because duplicate flattened names admit more than one topology. Removing the other affected designs leaves the tree-model ordering intact, although the original intent of those identifiers remains unknown.

The functional features depend on manual Boolean semantics because the pin-direction file supplies no module bodies. Scalar tests and Icarus cross-checks exercise the implemented equations. The zero-delay simulator still omits hazards and analog timing, and the three-sweep treatment of combinational cycles is an approximation. Structural-only runs avoid this uncertainty. These checks validate our implementation, not the timing behavior of a vendor library.

Hepp et al. [17] retain hierarchy and classify module graphs. Our flattened representation discards that unit, so the two studies answer different localization questions.

Seed sensitivity is limited but not absent. Activity seeds 7, 17, and 29 and model seeds 0-4 preserve the protocol order. The pinned environment reproduces the stored scores; it does not turn five related families into independent evidence.

### C. *Statistical conclusion validity*

The five LOFO folds are not independent experiments: each fit reuses four training families, and only five families are available. The bootstrap, jackknife, sign test, and seed runs therefore describe this corpus. They do not support a population-level claim about arbitrary circuits.

Trojan gates make up 0.69% of the sample, so a few false positives can move F1 sharply. AP, TPR, and FPR are reported beside F1 for that reason. Mixed and LOCO AP combine outputs from several fitted models; percentile normalization changes the LOCO magnitude more than the protocol ordering.

### D. *External validity*

The external boundary is narrow: 16 variants, five Trust-Hub families, one cell library, and fixed Trojans in flattened netlists. Systems-on-chip, other synthesis flows, layout-only modifications, side-channel detectors, and adaptive attackers are outside the experiment and should not be inferred from it.

IC fingerprinting [18] uses measurements from manufactured devices and can expose an anomalous implementation through physical signatures. This study has no such signal: every predictor comes from a flattened gate-level netlist.

ATTRITION [19] studies the attacker's side of evaluation. Its learning agent constructs Trojans intended to evade static detectors, whereas the Trust-Hub variants in this experiment are fixed before training begins. The present results do not cover an attacker adapting to the detector.

Recent node-wise graph learning [20] combines hardware-domain features with learned representations. Repeating the protocol comparison with that model would test whether a richer representation narrows the family-transfer gap reported here.

Independent circuit families and multiple synthesis flows are the next test. Until then, pooled, LOFO, and per-family rows should be read as a complete description of this sample, not as an estimate for industry-scale designs.

These omissions affect different predictors: new libraries change cell types and distances, new synthesis flows alter cones and levels, and adaptive attackers can target the features themselves. None appears in the five-family split.

## VIII. REPORTING IMPLICATIONS

A Trust-Hub audit is hard to reproduce from filenames alone. We used an explicit family manifest containing the distributed filename, host design, Trojan variant, known synthesis provenance, cell library, and parser exception. That file, rather than a prefix rule, defines family membership.

The three protocols can then be checked directly. A pooled split may mix related variants, LOCO tests one file from a familiar family, and LOFO removes the host logic. The five family rows reveal whether the headline reflects a broad pattern or one favorable host.

The written family rule fixes membership before fold construction and permits review without rerunning features.

We treated fold construction as an auditable object. Under LOCO, each circuit appears in exactly one test fold. Under LOFO, each family appears once and no host identifier crosses the split. Gate and positive counts remain in the manifest, because a formally valid split can still produce a small test set.

Reproducing Table IV requires the input hashes, family identifiers, gate-level fold assignments, and raw out-of-fold scores. The same record lists repaired or omitted netlists. Each prediction row carries the model, protocol, fold, circuit, family, gate instance, label, and score. Another group can then change the threshold or aggregation without retraining.

Model fitting remains separate from aggregation. Class weights come only from the training portion of a fold, and test labels enter after prediction. Gates are then grouped by family before the five family metrics are averaged. This order rules out full-dataset weighting and prevents averages that use incompatible units.

Keeping the raw scores also exposes score-scale shifts between folds. The s38584, s15850, and RS232 rows show why the macro value alone is insufficient.

Fold-level metric averages are not interchangeable across protocols because a fold denotes a gate subset, a circuit, or a family. Keeping gate-level rows avoids that ambiguity and lets the same predictions be re-aggregated at gate, circuit, and family levels.

We versioned the manifest and prediction file with the software environment. Any repair, exclusion, or family reassignment is recorded before feature extraction, which keeps the split auditable independently of model training.

A reader should be able to inspect the difficult s38584 transfer, the favorable s15850 result, and the small RS232 folds without inferring them from a macro average. Those cases show whether the headline is broad or carried by one benchmark.

Publishing the records permits later threshold and calibration studies to reuse the fitted classifiers and existing feature matrix.

None of these records requires a new detector. They are small by-products of the existing pipeline, but they preserve decisions that disappear from an aggregate table. Without them, a later comparison cannot tell whether a changed score comes from a model, a fold boundary, a parser repair, or an averaging rule.

## IX. CONCLUSION

Across 16 Trust-Hub netlists, RF and XGBoost lose both F1 and AP when a base family is held out. Feature, seed, parser, and reduced-sample checks preserve that direction. This is not a general claim about new hardware; it shows that sibling variants can make detector transfer look stronger than it is. Future Trust-Hub studies should release pooled and LOFO scores, five family rows, and the fold manifest. Those items are inexpensive to retain and expose whether the same host logic appears on both sides of a split. They keep benchmark reuse from being mistaken for a representation gain.

## REFERENCES

[1] M. Tehranipoor and F. Koushanfar, “A survey of hardware trojan taxonomy and detection,” IEEE Design & Test of Computers, vol. 27, no. 1, pp. 10-25, 2010, doi: 10.1109/MDT.2010.7.

[2] S. Bhunia, M. S. Hsiao, M. Banga, and S. Narasimhan, “Hardware trojan attacks: Threat analysis and countermeasures,” Proceedings of the IEEE, vol. 102, no. 8, pp. 1229-1247, 2014, doi: 10.1109/JPROC.2014.2334493.

[3] H. Salmani, “COTD: Reference-free hardware trojan detection and recovery based on controllability and observability in gate-level netlist,” IEEE Transactions on Information Forensics and Security, vol. 12, no. 2, pp. 338-350, 2017, doi: 10.1109/TIFS.2016.2613842.

[4] K. Hasegawa, M. Oya, M. Yanagisawa, and N. Togawa, “Hardware trojans classification for gate-level netlists based on machine learning,” in IEEE 22nd International Symposium on On-Line Testing and Robust System Design, 2016, pp. 203-206, doi: 10.1109/IOLTS.2016.7604700.

[5] C. Dong, J. Chen, W. Guo, and J. Zou, “A machine-learning-based hardware-trojan detection approach for chips in the internet of things,” International Journal of Distributed Sensor Networks, vol. 15, no. 12, 2019, art. no. 1550147719888098, doi: 10.1177/1550147719888098.

[6] H. Salmani, M. Tehranipoor, and R. Karri, “On design vulnerability analysis and trust benchmarks development,” in IEEE 31st International Conference on Computer Design, 2013, pp. 471-474, doi: 10.1109/ICCD.2013.6657085.

[7] B. Shakya, T. He, H. Salmani, D. Forte, S. Bhunia, and M. Tehranipoor, “Benchmarking of hardware trojans and maliciously affected circuits,” Journal of Hardware and Systems Security, vol. 1, no. 1, pp. 85-102, 2017, doi: 10.1007/s41635-017-0001-6.

[8] L. Breiman, “Random forests,” Machine Learning, vol. 45, no. 1, pp. 5-32, 2001, doi: 10.1023/A:1010933404324.

[9] T. Chen and C. Guestrin, “XGBoost: A scalable tree boosting system,” in Proceedings of the 22nd ACM SIGKDD International Conference on Knowledge Discovery and Data Mining, 2016, pp. 785-794, doi: 10.1145/2939672.2939785.

[10] F. Pedregosa, G. Varoquaux, A. Gramfort, V. Michel, B. Thirion, O. Grisel, M. Blondel, P. Prettenhofer, R. Weiss, V. Dubourg, J. VanderPlas, A. Passos, D. Cournapeau, M. Brucher, M. Perrot, and E. Duchesnay, “Scikit-learn: Machine learning in python,” Journal of Machine Learning Research, vol. 12, pp. 2825-2830, 2011.

[11] K. Hasegawa, M. Yanagisawa, and N. Togawa, “Trojan-feature extraction at gate-level netlists and its application to hardware-trojan detection using random forest classifier,” in IEEE International Symposium on Circuits and Systems, 2017, pp. 1-4, doi: 10.1109/ISCAS.2017.8050827.

[12] T. Kurihara and N. Togawa, “Hardware-trojan classification based on the structure of trigger circuits utilizing random forests,” in IEEE 27th International Symposium on On-Line Testing and Robust System Design, 2021, pp. 1-4, doi: 10.1109/IOLTS52814.2021.9486700.

[13] R. Negishi, T. Kurihara, and N. Togawa, “Hardware-trojan detection at gate-level netlists using a gradient boosting decision tree model and its extension using trojan probability propagation,” IEICE Transactions on Fundamentals of Electronics, Communications and Computer Sciences, vol. E107-A, no. 1, pp. 63-74, 2024, doi: 10.1587/transfun.2023KEP0005.

[14] D. Cheng, C. Dong, W. He, Z. Chen, and Y. Xu, “GNN4Gate: A bi-directional graph neural network for gate-level hardware trojan detection,” in Design, Automation & Test in Europe Conference, 2022, pp. 1315-1320.

[15] R. Yasaei, S.-Y. Yu, and M. A. A. Faruque, “GNN4TJ: Graph neural networks for hardware trojan detection at register transfer level,” in Design, Automation & Test in Europe Conference, 2021, pp. 1504-1509, doi: 10.23919/DATE51398.2021.9474174.

[16] H. Lashen, L. Alrahis, J. Knechtel, and O. Sinanoglu, “TrojanSAINT: Gate-level netlist sampling-based inductive learning for hardware trojan detection,” in IEEE International Symposium on Circuits and Systems, 2023, pp. 1-5, doi: 10.1109/ISCAS46773.2023.10181403.

[17] A. Hepp, J. Baehr, and G. Sigl, “Golden model-free hardware trojan detection by classification of netlist module graphs,” in Design, Automation & Test in Europe Conference, 2022, pp. 1317-1322, doi: 10.23919/DATE54114.2022.9774760.

[18] D. Agrawal, S. Baktir, D. Karakoyunlu, P. Rohatgi, and B. Sunar, “Trojan detection using IC fingerprinting,” in IEEE Symposium on Security and Privacy, 2007, pp. 296-310, doi: 10.1109/SP.2007.36.

[19] V. Gohil, H. Guo, S. Patnaik, and J. Rajendran, “ATTRITION: Attacking static hardware trojan detection techniques using reinforcement learning,” in Proceedings of the 2022 ACM SIGSAC Conference on Computer and Communications Security, 2022, pp. 1275-1289, doi: 10.1145/3548606.3560690.

[20] K. Hasegawa, K. Yamashita, S. Hidano, K. Fukushima, K. Hashimoto, and N. Togawa, “Node-wise hardware trojan detection based on graph learning,” IEEE Transactions on Computers, vol. 74, no. 3, pp. 749-761, 2025, doi: 10.1109/TC.2023.3280134.